\documentclass[sigconf,final]{acmart} 

\AtBeginDocument{%
  \providecommand\BibTeX{{%
    \normalfont B\kern-0.5em{\scshape i\kern-0.25em b}\kern-0.8em\TeX}}}

\setcopyright{acmcopyright}
\copyrightyear{2021}
\acmYear{2021}
\acmDOI{10.1145/1122445.1122456}

\acmConference[ICMI '21]{ICMI '21: 23rd ACM International Conference on Multimodal Interaction}{October 18--22, 2021}{Montreal, Canada}
\acmBooktitle{ICMI '21: 23rd ACM International Conference on Multimodal Interaction, Montreal, Canada}
\acmPrice{15.00}
\acmISBN{978-1-4503-XXXX-X/18/06}

\usepackage{array}
\usepackage{mdwmath}
\usepackage{mdwtab}
\usepackage{eqparbox}
\usepackage[utf8]{inputenc}
\usepackage{times}
\usepackage{latexsym}
\usepackage{times}
\usepackage{latexsym}
\usepackage{graphicx}
\usepackage{multirow}
\usepackage{subfig}
\usepackage{float}
\usepackage{algorithm}
\usepackage{algcompatible}
\usepackage{hyperref}
\usepackage[noend]{algpseudocode}
\usepackage{amsmath}
\usepackage{xcolor}
\usepackage{pgfplots}
\usepackage{tikz}
\usetikzlibrary{shapes,arrows}
\usetikzlibrary{positioning}
\usepackage{booktabs}
\usepackage{diagbox}
\usetikzlibrary{patterns}
\usepackage{soul,color}

\usepackage{url}
\usepackage{tabularx}

\usepackage{devanagari}

\begin{document}

\title{Toward Developing a Multimodal Multi-party Hindi Humorous Dataset for Humor Recognition in Conversations}

\title{M2H2: A Multimodal Multiparty Hindi Dataset For Humor Recognition in Conversations}

\author{\textbf{Dushyant Singh Chauhan$^\dagger$, Gopendra Vikram Singh$^\dagger$, Navonil Majumder$^+$, Amir Zadeh$^\mp$,} \\
\textbf{Asif Ekbal$^\dagger$, Pushpak Bhattacharyya$^\dagger$, Louis-philippe Morency$^\mp$, and Soujanya Poria$^+$}} 
\affiliation{%
  $^\dagger$ Department of Computer Science \& Engineering \\
  Indian Institute of Technology Patna \country{India} \\
  {\tt \{1821CS17, gopendra\_1921CS15,asif, pb\}@iitp.ac.in} \\
  $^\mp$ Language Technology Institute, Carnegie Mellon University, Pittsburgh \country{USA} \\
  {\tt \{abagherz, morency\}@cs.cmu.edu } \\
  $^+$ ISTD, Singapore University of Technology and Design \country{Singapore} \\
  {\tt \{navonil\_majumder, sporia\}@sutd.edu.sg }}

\ccsdesc[500]{Computing methodologies}
\ccsdesc[300]{Machine learning}
\ccsdesc{Machine learning approaches}
\ccsdesc[100]{Neural networks}

\keywords{Deep learning, Humor, Hindi dataset, Multimodal multiparty dataset}

\begin{abstract}

Humor recognition in conversations is a challenging task that has recently gained popularity due to its importance in dialogue understanding, including in multimodal settings (i.e., text, acoustics, and visual). The few existing datasets for humor are mostly in English. However, due to the tremendous growth in multilingual content, there is a great demand to build models and systems that support multilingual information access. To this end, we propose a dataset for \textit{Multimodal Multiparty Hindi Humor} (M2H2) recognition in conversations containing 6,191 utterances from 13 episodes of a very popular TV series \textit{"Shrimaan Shrimati Phir Se"}. Each utterance is annotated with humor/non-humor labels and encompasses acoustic, visual, and textual modalities. We propose several strong multimodal baselines and show the importance of contextual and multimodal information for humor recognition in conversations. The empirical results on \textit{M2H2} dataset demonstrate that multimodal information complements unimodal information for humor recognition. The dataset and the baselines are available at \url{http://www.iitp.ac.in/~ai-nlp-ml/resources.html} and \url{https://github.com/declare-lab/M2H2-dataset}.

\end{abstract}
\maketitle
\section{Introduction}\label{intro}



Humor \cite{warren2018humor} is defined as the nature of experiences to induce laughter and provide amusement. it can be implicit as well as explicit. Implicit humor may be expressed with neutral facial expressions and a polite tone, while explicit humor can be conveyed with laughter or funny facial expressions. Explicit humor may be easy to detect in comparison to implicit humor because of laughter or facial expression. So, while text \cite{bolkan2018humor,scheel2017definitions,boykoff2019laughing,khandelwal2018humor,sane2019deep,gal2019ironic,weller2019humor,blinov2019large,castro2017crowd,zhao2019embedding,garimella2020judge} alone may not always be enough to understand humor but if the utterance is multimodal in nature \cite{hasan2019ur,fallianda2018analyzing,ritschel2019irony,ritschel2020multimodal,mirnig2017elements,piata2020stylistic,song2021humor,vasquez2021cats,sabur2020identification,veronika2020multimodal,yang2019multimodal} and is accompanied with a video of the facial expressions and tone of the speaker, it can be easily understood that the utterance contains humor.

Humor recognition is especially challenging in many Indian languages because of the following reasons: unavailability of resources and tools; morphological richness; free word order etc. In our current work, we focus on Hindi, one of the most popularly spoken languages in India. In terms of the number of speakers, Hindi ranks third\footnote{\url{https://currentaffairs.adda247.com/hindi-ranks-3rd-most-spoken-language-in-the-world/}} all over in the world. In comparison to English, Hindi is much richer in terms of morphology. In Hindi, all the nouns and adjectives have genders, and the verb agrees in number and gender with them. While, English is free of that confounding factor of the gender of a common noun, and is also free of the further complication that the verb has to agree with the gender of the noun or pronoun governing it. These challenges motivated us to do some qualitative research work for Humor detection in Hindi.

In Hindi, there is no available dataset for humour detection (neither unimodal nor multimodal). In our work, we introduce a new dataset, named as M2H2, which includes not only textual dialogues but also their corresponding visual and audio counterparts. The main contributions 
of our proposed research are as follows: 
\begin{itemize}
 \item We propose a dataset for \textit{Multimodal Multi-party Hindi Humor} recognition in conversations. There are 6,191 utterances in the M2H2 dataset;
 \item We set up two strong baselines, \textit{viz.}, DialogueRNN~\cite{majumder2019dialoguernn} and bcLSTM~\cite{poria2017context} with MISA~\cite{hazarika2020misa} for multimodal fusion. Both DialogueRNN and bcLSTM are contextual models.
 \item Empirical results on the \textit{M2H2} dataset show the efficacy of multimodal information over only textual information for humor detection.
\end{itemize}


    
    
\section{Dataset}\label{dataset}


\begin{figure*}[t]
\begin{center}
\includegraphics[width=0.9\textwidth]{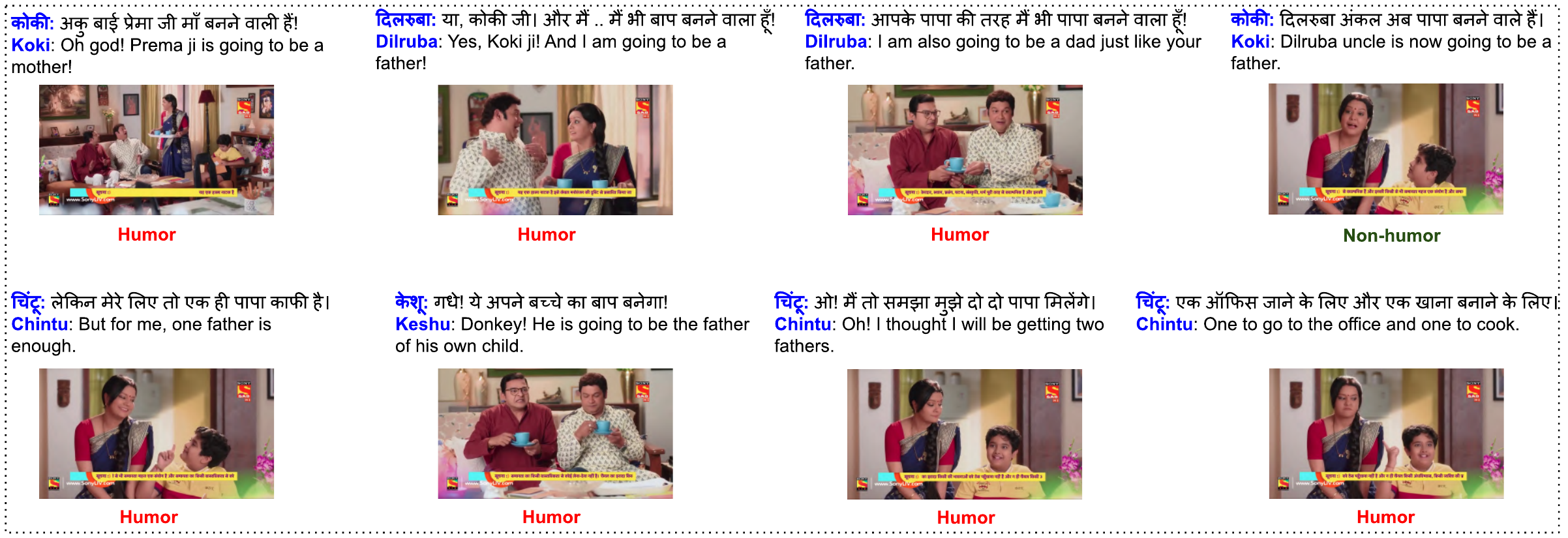}   
\caption{Some samples from dataset.}
\label{fig:samples_dataset}
\end{center} 
\end{figure*}


We gathered samples from a famous TV show \textit{"Shrimaan Shrimati Phir Se"}\footnote{Total of 4.46 hours in length} and annotated them manually. We make groups of these samples (utterances) based on their context into scenes. Each utterance in each scene consists of a label indicating humor of that utterance i.e., humor\footnote{There are 1/$3^{rd}$ utterances which are humorous.} or non-humor. Besides, each utterance is also annotated with its speaker and listener information. In multiparty conversation, listener identification poses a great challenge. In our dataset, we define the listener as the party in the conversation to whom the speaker is replying. Each utterance in each scene is coupled with its context utterances, which are preceding turns by the speakers participating in the conversation. It also contains multi-party conversations that are more challenging to classify than dyadic variants. We show the dataset statistics in Table \ref{tab:dataset_statistics}.

\begin{table}[!h]
\centering
\small
\resizebox{0.3\textwidth}{!}
{
\begin{tabular}{|l|c|}
\hline
\bf Description  & \bf Statistic \\ \hline
\textit{\#Scenes} & 136 \\ 

\textit{\#Utterances} & 6191 \\ 
\textit{Min \#Utterances per Scene} & 8 \\ 
\textit{Max \#Utterances per Scene} & 155 \\  
\textit{\#Utterances per Scene} & 45.52 \\ 
\textit{\#Words} & 34472 \\ 
\textit{\#Unique words} & 5454 \\ 


\textit{\#Humor class} & 2089 \\ 
\textit{\#Non-humor class} & 4102 \\ 

\textit{\#Speakers} & 41 \\
\textit{Total duration} & 4.46 Hours \\

\hline
\end{tabular}
}
\caption{Dataset statistics.}
\label{tab:dataset_statistics}
\end{table}



\subsection{Challenges}
We crawled the video file and transcription file using \textit{YouTube} for the TV show \textit{"Shrimaan Shrimati Phir Se"}. The downloaded Hindi transcription file had some errors. So, we hired transcribers, who were experts in the Hindi language. They wrote Hindi utterances based on the audio files. 

As the context is based on the scenes, we divide the whole dataset into scenes based on its context. Each utterance in each scene consists of a humor label (i.e., humor or non-humor) and speaker and listener information. 


\subsection{Annotation Guidelines}
We employ three Ph.D. students with high proficiency in Hindi and with prior experience in labeling \textit{Humor} and \textit{Non-humor} in conversational settings.

The guidelines for annotation, along with some examples (c.f. Figure \ref{fig:samples_dataset}), were explained to the annotators before starting the annotation process. The annotators were asked to annotate every utterance with humor/non-humor and corresponding speaker and listener information. We also annotated 20 annotations by ourselves and set it as a quality checker to evaluate the quality of the annotators. A majority voting scheme was used for selecting the final humor/non-humor. We achieve an overall Fleiss' \cite{fleiss1971measuring} kappa score of 0.84, which is considered to be reliable.


\section{Baseline Methods}\label{proposed}

In this section, we provide strong benchmarks for \textit{M2H2} dataset. We employ three strong baselines frameworks for humor detection. 


\subsection{Strong Baseline for Multimodal Fusion: MISA}
\textit{MISA} (Modality-Invariant and-Specific Representations for Multimodal Sentiment Analysis) \cite{hazarika2020misa} has two main stages: modality representation learning and modality fusion. It divides each modality into two sub-spaces. The first subspace is modality-invariant, in which representations from different modalities learn to share similarities and minimize the distance between them. The second subspace is modality-specific, which is unique to every modality and contains its distinguishing characteristics. These representations give a comprehensive perspective of multimodal data, which is utilized for fusion and task prediction. The model shows the state-of-the-art (SOTA) for MOSI \cite{zadeh:mosi} and MOSEI \cite{zadeh2018acl} datasets.

\subsection{Strong Conversation Classification Baseline \#1: DialogueRNN}
DialogueRNN \cite{majumder2019dialoguernn} is a multi-party framework tailored for modelling emotions and sentiments in conversations. In DialogueRNN, they explained a novel recurrent neural network-based system that tracks of each party state during a discussion and uses this knowledge to classify emotions. It employs three levels of gated recurrent units (GRU) to represent the conversational context in order to accurately recognize emotions, intensity, and attitudes in a discussion. They showed the SOTA performance on two different datasets i.e., IEMOCAP \cite{busso2008iemocap} and AVEC \cite{schuller2012avec}.  

\subsection{Strong Conversation Classification Baseline \#2: bcLSTM} 
bcLstm \cite{poria2017context} is a bidirectional contextual LSTM. Bi-directional LSTMs are formed by stacking two uni-directional LSTMs with opposing directions. As a result, an utterance can learn from utterances that come before and after it in the video which is of course context. 

Leveraging the above three strong baselines, we formulate two setups, \textit{viz.} MISA with DialogueRNN (MISA+DialogueRNN) and MISA with bcLSTM (MISA+bcLSTM). Here, MISA acts as a fusion model while DialogueRNN and bcLSTM are contextual models for conversation classification. DialogueRNN and bcLSTM have shown excellent performance in different conversation classification tasks such as Emotion Recognition Conversation. For MISA+DialogueRNN, we first pass multimodal inputs through MISA and obtain the fused features. These fused features are then passed through DialogueRNN for humor classification. Similarly, for MISA+bcLSTM, We first pass multimodal inputs through MISA and obtain the fused features. Then, these fused features are passed through bcLSTM for humor classification.



\subsection{Feature Extraction}

For textual features, we take the pre-trained 300-dimensional Hindi \textit{fastText} embeddings \cite{joulin2016fasttext}. For visual feature extraction, we use 3D-ResNeXt-101\footnote{\url{https://github.com/kaiqiangh/extracting-video-features-ResNeXt}} \cite{xie2017aggregated} which is pre-trained on Kinetics at a rate of 1.5 features per second and a resolution of 112. While, we use openSMILE \cite{eyben2010opensmile} for acoustic feature extraction. openSMILE\footnote{\url{https://github.com/audeering/opensmile}} can extract Low-Level Descriptors (LLD) and change them using different filters, functions, and transformations. We use a tonal low-level features group of openSMILE to extract the features.

\section{Experiments and Analysis}\label{experiments}

\subsection{Experimental Setup}
We evaluate our proposed model on the \textit{M2H2} dataset. We perform five-fold cross-validation for experiments. Empirically, We take five\footnote{Baseline models give the best result at five.} utterances as context for a particular utterance. 



We implement our proposed model on the Python-based PyTorch\footnote{\url{https://pytorch.org/}} deep learning library. As the evaluation metric, we employ precision (P), recall (R), and F1-score (F1) for humor recognition. We use \textit{Adam} as an optimizer, \textit{Softmax} as a classifier for humor detection, and the \textit{categorical cross-entropy} as a loss function.

\subsection{Results and Analysis}

We evaluate our proposed architecture with all the possible input combinations (c.f. Table \ref{tab:M_DRNN_bcLSTM}) i.e. unimodal \textit{(T, A, V)}, bimodal (\textit{T+V, T+A, A+V}) and trimodal (\textit{T+V+A}). 


For MISA+DialogueRNN, trimodal achieves the best precision of 71.21\% ($16.05$\footnote{\textit{maximum improvement over unimodal (V)}} points $\uparrow$ and $9.51$\footnote{\textit{maximum improvement over bimodal (A+V)}} points $\uparrow$), recall of 72.11\% ($14.79$ points $\uparrow$ and $8.9$ points $\uparrow$) and F1-score of 71.67\% ($14.93$ points $\uparrow$ and $9.23$ points $\uparrow$). We observe that trimodal performs better than the unimodal and bimodal. We show the results in Table \ref{tab:M_DRNN_bcLSTM}.

\begin{table}[t]
\small
\centering
\resizebox{0.49\textwidth}{!}
{    
\begin{tabular}{|c|c|c|c||c|c|c|}
\hline

 & \multicolumn{3}{|c||}{\bf \em  MISA+DialogueRNN} & \multicolumn{3}{|c|}{\bf \em  MISA+bcLSTM} \\ \cline{2-7}

\bf \em Labels & \bf P & \bf R & \bf F1  & \bf P & \bf R & \bf F1  \\ \hline \hline

\bf \em T    & 67.11 & 68.21 & 67.65 & 66.27 & 66.51 & 66.38\\
\bf \em A    & 57.91 & 59.10 & 58.52 & 57.51 & 58.92 & 58.21    \\
\bf \em V    & 55.16 & 57.32 & 56.74 & 53.13 & 55.21 & 54.19  \\ \hline
\bf \em T+V  & 70.03 & 70.61 & 70.31 & 67.74 & 68.89 & 68.31 \\
\bf \em T+A  & 69.70 & 69.90 & 69.63 & 67.41 & 68.12 & 67.75  \\
\bf \em A+V  & 61.70 & 63.21 & 62.44 & 59.61 & 61.23 & 60.40 \\ \hline
\bf \em T+V+A & \bf 71.21 & \bf 72.11 & \bf 71.67 & \bf  69.04 & \bf  69.83& \bf 69.43  \\

\hline
\end{tabular}
}
\caption{Experiment results for MISA+DialogueRNN and MISA+bcLSTM.}
\label{tab:M_DRNN_bcLSTM}
\end{table}

Similarly, for MISA+bcLSTM, trimodal achieves the best precision of 69.04\% ($15.91$\footnote{\textit{maximum improvement over unimodal (V)}} points $\uparrow$ and $9.43$\footnote{\textit{maximum improvement over bimodal (A+V)}} points $\uparrow$), recall of 69.83\% ($14.62$ points $\uparrow$ and $8.6$ points $\uparrow$) and F1-score of 69.43\% ($15.24$ points $\uparrow$ and $9.03$ points $\uparrow$). We observe that trimodal performs better than the unimodal and bimodal. 

\begin{table*}[ht!]
\small
\centering
\resizebox{0.9\textwidth}{!}
{
\begin{tabular}{|l|p{23em}|p{23em}||c||c|c| }
\hline

\multicolumn{6}{|c|}{\bf \em Correct Prediction} \\ \hline
& & & & \multicolumn{2}{c|}{\bf \em Predicted} \\ \cline{4-6}

& \bf  & \bf & & \bf Unimodal & \bf Multimodal \\ 
& \bf Hindi Utterances & \bf English Utterances & \bf Actual & \bf (T) & \bf (T+V+A) \\ \hline \hline

1 & {\dn ar\? \7{b}YAp\? m\?{\qva} m\4{\qva} \7{t}Mh\?{\qva} \7{K}d l\? k\? jAtA.}
& Oh in old age I would have taken you myself. & humor & \textcolor{red}{non-humor} & humor \\ \hline

2 & {\dn y\? EdlzbA khtA h\4 nA Ek EpCl\? j\306wm m\?{\qva} vo Co\8{V}ml mo\8{V}ml kroXpEt kA eklOtA b\?VA TA{\rs ,\re} rAiV{\rs ?\re}}
& This Dilruba says that he was the only son of Chhotumal Motumal Crorepati in his previous life, right? & non-humor & \textcolor{red}{humor} & non-humor \\ \hline

3 & {\dn -VAP vAlo{\qva}{\rs !\re} m\?rA mtlb}
& Staff guys! I mean & humor & \textcolor{red}{non-humor} & humor \\ \hline

4 & {\dn \7{k}C yAd aAyA{\rs ?\re} \7{k}C yAd aAyA{\rs ?\re}}
& Remember anything? remember something? & non-humor & \textcolor{red}{humor} & non-humor \\ \hline

5 & {\dn jAt\? v\3C4w e\?sF a\7{f}B bAt\?{\qva} nhF{\qva} krt\?.}
& Do not do such inauspicious things while leaving. & humor & \textcolor{red}{non-humor} & humor  \\ \hline \hline

\multicolumn{6}{|c|}{\bf \em Incorrect Prediction} \\ \hline

1 & {\dn y\? log \7{m}J\? b\7{h}t mAr rh\? h\4{\qva}.}
& These guys are beating me hard & humor & \textcolor{red}{non-humor} & \textcolor{red}{non-humor} \\ \hline

2 & {\dn hAy{\rs ,\re} VoVo{\rs !\re}}
& hey, Toto! & non-humor & non-humor & \textcolor{red}{humor} \\ \hline

3 & {\dn y\? Ep\2kF lAI.}
& Pinky brought this & humor & humor & \textcolor{red}{non-humor} \\ \hline

4 & {\dn ar\? BAI{\rs ,\re} m\?rF aslF vAlF gn yhF{\qva} \8{C}V g\4i.}
& Hey brother, my original gun is left here & non-humor & \textcolor{red}{humor} & \textcolor{red}{humor} \\ \hline

5 & {\dn kOn sF aslF h\4 kOn sF nklF h\4.}
& which is real which is fake & non-humor & non-humor & \textcolor{red}{humor}  \\ \hline

\end{tabular}
}
\caption{Error analysis: Some correct and incorrect predicted samples.}
\label{tab-error_analysis}
\end{table*}

\subsection{Ablation Study}


We also perform an ablation study (c.f. Table \ref{tab:ablation}) to show the efficacy of contextual models. We perform experiments on MISA (i.e., without contextual information). As per the result, we can see when MISA is used with contextual models (bcLSTM or DialogueRNN) it shows significant improvement rather than when it uses alone.

\begin{table}[!h]
\centering
\resizebox{0.3\textwidth}{!}
{    
\begin{tabular}{|l|c|c|c| }
\hline

\bf \em Setup &  \bf P & \bf R & \bf F1  \\ \hline \hline

\bf \em MISA+DialogueRNN & 71.21  & 72.11 & 71.67 \\ \hline 
\bf \em MISA+bcLSTM & 69.04 & 69.83 & 69.43 \\ \hline 
\bf \em MISA & 67.61 & 70.26 & 68.90 \\ \hline

\end{tabular}
}
\caption{Ablation study}
\label{tab:ablation}
\end{table}

\subsection{Error Analysis}\label{error}

In this section, we perform error analysis for our baseline system i.e, MISA+DialogueRNN. We take some samples which are correctly and incorrectly predicted by the baseline model to analyze the model's strengths and weaknesses.

\begin{figure}[!h]
\begin{center}
\includegraphics[width=0.49\textwidth]{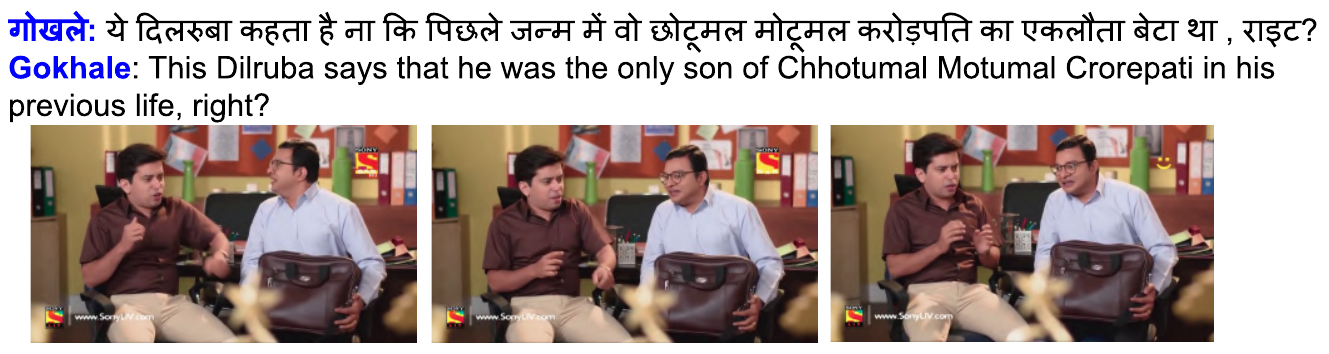} 
\caption{Some visual frames of second utterance of correct prediction in Table \ref{tab-error_analysis}.}
\label{fig:corrected_visual_frames}
\end{center} 
\end{figure}

We show a few samples\footnote{For the global audience, we also translate these Hindi utterances into English.} (c.f. Table \ref{tab-error_analysis}) which are correctly predicted by the multimodal setup, but incorrectly predicted by the unimodal setup. For example, second utterance of the corrected prediction in Table \ref{tab-error_analysis}, 
{\dn y\? EdlzbA khtA h\4 nA Ek EpCl\? j\306wm m\?{\qva} vo Co\8{V}ml mo\8{V}ml kroXpEt kA eklOtA b\?VA TA{\rs ,\re} rAiV{\rs ?\re}} (This Dilruba says that he was the only son of Chhotumal Motumal Crorepati in his previous life, right?) Gokhale is giving idea to Keshav with a polite tone and neutral facial expression. The actual label is non-humor but the unimodal model (text) predicts it as humorous instance while the multimodal model predicts the label correctly because it is clear from the acoustic and visual features that there were no humorous punches but only normal talk. We show some visual frames in Figure~\ref{fig:corrected_visual_frames} which clearly show the neutral facial expression.


\begin{figure}[!h]
\begin{center}
\includegraphics[width=0.49\textwidth]{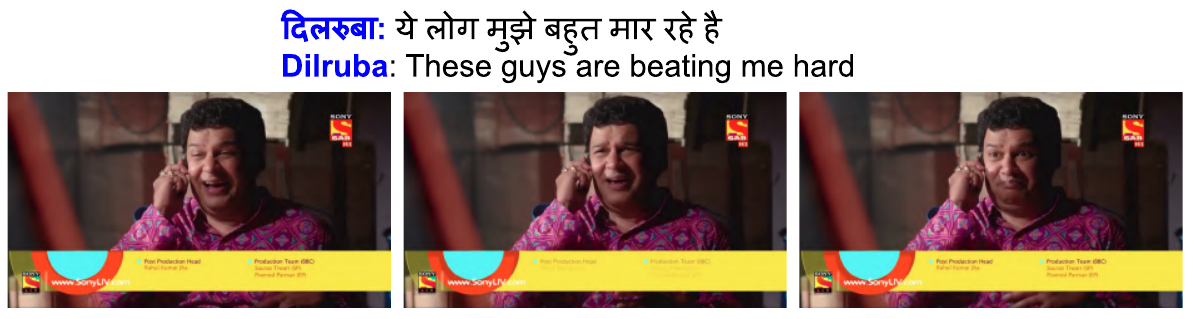}
\caption{Some visual frames of first utterance of incorrect prediction in Table \ref{tab-error_analysis}.}
\label{fig:incorrected_visual_frames}
\end{center} 
\end{figure}

On the other hand, we also show some samples, where the multimodal model fails to predict the correct label. For example, first utterance of incorrect prediction in Table \ref{tab-error_analysis}, {\dn y\? log \7{m}J\? b\7{h}t mAr rh\? h\4{\qva}.} (these guys are beating me hard) has the actual label as humor. But, the multimodal fails to predict the correct label because of the sad tone and the sad visual expression of Dilruba ({\dn EdlzbA}). We show some visual frames in Figure~\ref{fig:incorrected_visual_frames} which clearly shows the sad facial expression of Dilruba.

\section{Conclusion}\label{conclusion}

In this paper, we have presented a novel \textit{Multimodal Multi-party Hindi Humor} (M2H2) recognition dataset for conversations (4.46 hours in length). 
We employed two strong baseline setups, \textit{viz.} MISA with DialogueRNN and MISA with bcLSTM. MISA is a fusion model while DialogueRNN and bcLSTM are contextual models. 
We believe this dataset will also be useful as a useful resource for both conversational humor recognition and multimodal artificial intelligence. Empirical results on the \textit{M2H2} dataset demonstrate that the multimodal baselines yield better performance over the unimodal framework.

In the future, we would like to extend our work towards the multi-party dialogue generation in Hindi with the help of humor and speaker information.

\section*{Acknowledgement}

Dushyant Singh Chauhan acknowledges the support of Prime Minister Research Fellowship (PMRF), Govt. of India. Asif Ekbal acknowledges the Young Faculty Research Fellowship (YFRF), supported by Visvesvaraya
PhD scheme for Electronics and IT, Ministry of Electronics and Information Technology (Meit/8Y), Government of India, being implemented by Digital India Corporation (formerly Media Lab Asia). Soujanya Poria is supported by the SUTD SRG grant no. T1SRIS19149.

\bibliographystyle{ACM-Reference-Format}
\bibliography{bib/Reference,bib/bib1,bib/bib2,bib/bib3,bib/bib4}


\begin{thebibliography}{34}


\ifx \showCODEN    \undefined \def \showCODEN     #1{\unskip}     \fi
\ifx \showDOI      \undefined \def \showDOI       #1{#1}\fi
\ifx \showISBNx    \undefined \def \showISBNx     #1{\unskip}     \fi
\ifx \showISBNxiii \undefined \def \showISBNxiii  #1{\unskip}     \fi
\ifx \showISSN     \undefined \def \showISSN      #1{\unskip}     \fi
\ifx \showLCCN     \undefined \def \showLCCN      #1{\unskip}     \fi
\ifx \shownote     \undefined \def \shownote      #1{#1}          \fi
\ifx \showarticletitle \undefined \def \showarticletitle #1{#1}   \fi
\ifx \showURL      \undefined \def \showURL       {\relax}        \fi
\providecommand\bibfield[2]{#2}
\providecommand\bibinfo[2]{#2}
\providecommand\natexlab[1]{#1}
\providecommand\showeprint[2][]{arXiv:#2}

\bibitem[\protect\citeauthoryear{Blinov, Bolotova-Baranova, and
  Braslavski}{Blinov et~al\mbox{.}}{2019}]%
        {blinov2019large}
\bibfield{author}{\bibinfo{person}{Vladislav Blinov}, \bibinfo{person}{Valeria
  Bolotova-Baranova}, {and} \bibinfo{person}{Pavel Braslavski}.}
  \bibinfo{year}{2019}\natexlab{}.
\newblock \showarticletitle{Large dataset and language model fun-tuning for
  humor recognition}. In \bibinfo{booktitle}{\emph{Proceedings of the 57th
  annual meeting of the association for computational linguistics}}.
  \bibinfo{pages}{4027--4032}.
\newblock


\bibitem[\protect\citeauthoryear{Bolkan, Griffin, and Goodboy}{Bolkan
  et~al\mbox{.}}{2018}]%
        {bolkan2018humor}
\bibfield{author}{\bibinfo{person}{San Bolkan}, \bibinfo{person}{Darrin~J
  Griffin}, {and} \bibinfo{person}{Alan~K Goodboy}.}
  \bibinfo{year}{2018}\natexlab{}.
\newblock \showarticletitle{Humor in the classroom: the effects of integrated
  humor on student learning}.
\newblock \bibinfo{journal}{\emph{Communication Education}}
  \bibinfo{volume}{67}, \bibinfo{number}{2} (\bibinfo{year}{2018}),
  \bibinfo{pages}{144--164}.
\newblock


\bibitem[\protect\citeauthoryear{Boykoff and Osnes}{Boykoff and Osnes}{2019}]%
        {boykoff2019laughing}
\bibfield{author}{\bibinfo{person}{Maxwell Boykoff} {and} \bibinfo{person}{Beth
  Osnes}.} \bibinfo{year}{2019}\natexlab{}.
\newblock \showarticletitle{A laughing matter? Confronting climate change
  through humor}.
\newblock \bibinfo{journal}{\emph{Political Geography}}  \bibinfo{volume}{68}
  (\bibinfo{year}{2019}), \bibinfo{pages}{154--163}.
\newblock


\bibitem[\protect\citeauthoryear{Busso, Bulut, Lee, Kazemzadeh, Mower, Kim,
  Chang, Lee, and Narayanan}{Busso et~al\mbox{.}}{2008}]%
        {busso2008iemocap}
\bibfield{author}{\bibinfo{person}{Carlos Busso}, \bibinfo{person}{Murtaza
  Bulut}, \bibinfo{person}{Chi-Chun Lee}, \bibinfo{person}{Abe Kazemzadeh},
  \bibinfo{person}{Emily Mower}, \bibinfo{person}{Samuel Kim},
  \bibinfo{person}{Jeannette~N Chang}, \bibinfo{person}{Sungbok Lee}, {and}
  \bibinfo{person}{Shrikanth~S Narayanan}.} \bibinfo{year}{2008}\natexlab{}.
\newblock \showarticletitle{IEMOCAP: Interactive emotional dyadic motion
  capture database}.
\newblock \bibinfo{journal}{\emph{Language resources and evaluation}}
  \bibinfo{volume}{42}, \bibinfo{number}{4} (\bibinfo{year}{2008}),
  \bibinfo{pages}{335}.
\newblock


\bibitem[\protect\citeauthoryear{Castro, Chiruzzo, Ros{\'a}, Garat, and
  Moncecchi}{Castro et~al\mbox{.}}{2017}]%
        {castro2017crowd}
\bibfield{author}{\bibinfo{person}{Santiago Castro}, \bibinfo{person}{Luis
  Chiruzzo}, \bibinfo{person}{Aiala Ros{\'a}}, \bibinfo{person}{Diego Garat},
  {and} \bibinfo{person}{Guillermo Moncecchi}.}
  \bibinfo{year}{2017}\natexlab{}.
\newblock \showarticletitle{A crowd-annotated spanish corpus for humor
  analysis}.
\newblock \bibinfo{journal}{\emph{arXiv preprint arXiv:1710.00477}}
  (\bibinfo{year}{2017}).
\newblock


\bibitem[\protect\citeauthoryear{Eyben, W{\"o}llmer, and Schuller}{Eyben
  et~al\mbox{.}}{2010}]%
        {eyben2010opensmile}
\bibfield{author}{\bibinfo{person}{Florian Eyben}, \bibinfo{person}{Martin
  W{\"o}llmer}, {and} \bibinfo{person}{Bj{\"o}rn Schuller}.}
  \bibinfo{year}{2010}\natexlab{}.
\newblock \showarticletitle{Opensmile: the munich versatile and fast
  open-source audio feature extractor}. In
  \bibinfo{booktitle}{\emph{Proceedings of the 18th ACM international
  conference on Multimedia}}. ACM, \bibinfo{pages}{1459--1462}.
\newblock


\bibitem[\protect\citeauthoryear{Fallianda, Astiti, and Hanim}{Fallianda
  et~al\mbox{.}}{2018}]%
        {fallianda2018analyzing}
\bibfield{author}{\bibinfo{person}{Fallianda Fallianda},
  \bibinfo{person}{Rani~Yuni Astiti}, {and} \bibinfo{person}{Zulvy~Alivia
  Hanim}.} \bibinfo{year}{2018}\natexlab{}.
\newblock \showarticletitle{Analyzing Humor in Newspaper Comic Strips Using
  Verbal-Visual Analysis}.
\newblock \bibinfo{journal}{\emph{Lingua Cultura}} \bibinfo{volume}{12},
  \bibinfo{number}{4} (\bibinfo{year}{2018}), \bibinfo{pages}{383--388}.
\newblock


\bibitem[\protect\citeauthoryear{Fleiss}{Fleiss}{1971}]%
        {fleiss1971measuring}
\bibfield{author}{\bibinfo{person}{Joseph~L. Fleiss}.}
  \bibinfo{year}{1971}\natexlab{}.
\newblock \showarticletitle{Measuring nominal scale agreement among many
  rater}.
\newblock \bibinfo{journal}{\emph{Psychological Bulletin}}
  \bibinfo{volume}{76} (\bibinfo{year}{1971}), \bibinfo{pages}{378--382}.
\newblock


\bibitem[\protect\citeauthoryear{Gal}{Gal}{2019}]%
        {gal2019ironic}
\bibfield{author}{\bibinfo{person}{Noam Gal}.} \bibinfo{year}{2019}\natexlab{}.
\newblock \showarticletitle{Ironic humor on social media as participatory
  boundary work}.
\newblock \bibinfo{journal}{\emph{New Media \& Society}} \bibinfo{volume}{21},
  \bibinfo{number}{3} (\bibinfo{year}{2019}), \bibinfo{pages}{729--749}.
\newblock


\bibitem[\protect\citeauthoryear{Garimella, Banea, Hossain, and
  Mihalcea}{Garimella et~al\mbox{.}}{2020}]%
        {garimella2020judge}
\bibfield{author}{\bibinfo{person}{Aparna Garimella}, \bibinfo{person}{Carmen
  Banea}, \bibinfo{person}{Nabil Hossain}, {and} \bibinfo{person}{Rada
  Mihalcea}.} \bibinfo{year}{2020}\natexlab{}.
\newblock \showarticletitle{" Judge me by my size (noun), do you?''YodaLib: A
  Demographic-Aware Humor Generation Framework}.
\newblock \bibinfo{journal}{\emph{arXiv preprint arXiv:2006.00578}}
  (\bibinfo{year}{2020}).
\newblock


\bibitem[\protect\citeauthoryear{Hasan, Rahman, Zadeh, Zhong, Tanveer, Morency,
  et~al\mbox{.}}{Hasan et~al\mbox{.}}{2019}]%
        {hasan2019ur}
\bibfield{author}{\bibinfo{person}{Md~Kamrul Hasan}, \bibinfo{person}{Wasifur
  Rahman}, \bibinfo{person}{Amir Zadeh}, \bibinfo{person}{Jianyuan Zhong},
  \bibinfo{person}{Md~Iftekhar Tanveer}, \bibinfo{person}{Louis-Philippe
  Morency}, {et~al\mbox{.}}} \bibinfo{year}{2019}\natexlab{}.
\newblock \showarticletitle{Ur-funny: A multimodal language dataset for
  understanding humor}.
\newblock \bibinfo{journal}{\emph{arXiv preprint arXiv:1904.06618}}
  (\bibinfo{year}{2019}).
\newblock


\bibitem[\protect\citeauthoryear{Hazarika, Zimmermann, and Poria}{Hazarika
  et~al\mbox{.}}{2020}]%
        {hazarika2020misa}
\bibfield{author}{\bibinfo{person}{Devamanyu Hazarika}, \bibinfo{person}{Roger
  Zimmermann}, {and} \bibinfo{person}{Soujanya Poria}.}
  \bibinfo{year}{2020}\natexlab{}.
\newblock \showarticletitle{MISA: Modality-Invariant and-Specific
  Representations for Multimodal Sentiment Analysis}.
\newblock \bibinfo{journal}{\emph{arXiv preprint arXiv:2005.03545}}
  (\bibinfo{year}{2020}).
\newblock


\bibitem[\protect\citeauthoryear{Joulin, Grave, Bojanowski, Douze, J{\'e}gou,
  and Mikolov}{Joulin et~al\mbox{.}}{2016}]%
        {joulin2016fasttext}
\bibfield{author}{\bibinfo{person}{Armand Joulin}, \bibinfo{person}{Edouard
  Grave}, \bibinfo{person}{Piotr Bojanowski}, \bibinfo{person}{Matthijs Douze},
  \bibinfo{person}{H{\'e}rve J{\'e}gou}, {and} \bibinfo{person}{Tomas
  Mikolov}.} \bibinfo{year}{2016}\natexlab{}.
\newblock \showarticletitle{FastText.zip: Compressing text classification
  models}.
\newblock \bibinfo{journal}{\emph{arXiv preprint arXiv:1612.03651}}
  (\bibinfo{year}{2016}).
\newblock


\bibitem[\protect\citeauthoryear{Khandelwal, Swami, Akhtar, and
  Shrivastava}{Khandelwal et~al\mbox{.}}{2018}]%
        {khandelwal2018humor}
\bibfield{author}{\bibinfo{person}{Ankush Khandelwal}, \bibinfo{person}{Sahil
  Swami}, \bibinfo{person}{Syed~S Akhtar}, {and} \bibinfo{person}{Manish
  Shrivastava}.} \bibinfo{year}{2018}\natexlab{}.
\newblock \showarticletitle{Humor detection in english-hindi code-mixed social
  media content: Corpus and baseline system}.
\newblock \bibinfo{journal}{\emph{arXiv preprint arXiv:1806.05513}}
  (\bibinfo{year}{2018}).
\newblock


\bibitem[\protect\citeauthoryear{Majumder, Poria, Hazarika, Mihalcea, Gelbukh,
  and Cambria}{Majumder et~al\mbox{.}}{2019}]%
        {majumder2019dialoguernn}
\bibfield{author}{\bibinfo{person}{Navonil Majumder}, \bibinfo{person}{Soujanya
  Poria}, \bibinfo{person}{Devamanyu Hazarika}, \bibinfo{person}{Rada
  Mihalcea}, \bibinfo{person}{Alexander Gelbukh}, {and} \bibinfo{person}{Erik
  Cambria}.} \bibinfo{year}{2019}\natexlab{}.
\newblock \showarticletitle{DialogueRNN: An Attentive {RNN} for Emotion
  Detection in Conversations}. In \bibinfo{booktitle}{\emph{The Thirty-Third
  {AAAI} Conference on Artificial Intelligence, {AAAI} 2019, The Thirty-First
  Innovative Applications of Artificial Intelligence Conference, {IAAI} 2019,
  The Ninth {AAAI} Symposium on Educational Advances in Artificial
  Intelligence, {EAAI} 2019, Honolulu, Hawaii, USA, January 27 - February 1,
  2019}}, Vol.~\bibinfo{volume}{33}. \bibinfo{pages}{6818--6825}.
\newblock


\bibitem[\protect\citeauthoryear{Mirnig, Stollnberger, Giuliani, and
  Tscheligi}{Mirnig et~al\mbox{.}}{2017}]%
        {mirnig2017elements}
\bibfield{author}{\bibinfo{person}{Nicole Mirnig}, \bibinfo{person}{Gerald
  Stollnberger}, \bibinfo{person}{Manuel Giuliani}, {and}
  \bibinfo{person}{Manfred Tscheligi}.} \bibinfo{year}{2017}\natexlab{}.
\newblock \showarticletitle{Elements of humor: How humans perceive verbal and
  non-verbal aspects of humorous robot behavior}. In
  \bibinfo{booktitle}{\emph{Proceedings of the Companion of the 2017 ACM/IEEE
  International Conference on Human-Robot Interaction}}.
  \bibinfo{pages}{211--212}.
\newblock


\bibitem[\protect\citeauthoryear{Piata}{Piata}{2020}]%
        {piata2020stylistic}
\bibfield{author}{\bibinfo{person}{Anna Piata}.}
  \bibinfo{year}{2020}\natexlab{}.
\newblock \showarticletitle{Stylistic humor across modalities: The case of
  Classical Art Memes}.
\newblock \bibinfo{journal}{\emph{Internet Pragmatics}} \bibinfo{volume}{3},
  \bibinfo{number}{2} (\bibinfo{year}{2020}), \bibinfo{pages}{174--201}.
\newblock


\bibitem[\protect\citeauthoryear{Poria, Cambria, Hazarika, Majumder, Zadeh, and
  Morency}{Poria et~al\mbox{.}}{2017}]%
        {poria2017context}
\bibfield{author}{\bibinfo{person}{Soujanya Poria}, \bibinfo{person}{Erik
  Cambria}, \bibinfo{person}{Devamanyu Hazarika}, \bibinfo{person}{Navonil
  Majumder}, \bibinfo{person}{Amir Zadeh}, {and}
  \bibinfo{person}{Louis-Philippe Morency}.} \bibinfo{year}{2017}\natexlab{}.
\newblock \showarticletitle{Context-dependent sentiment analysis in
  user-generated videos}. In \bibinfo{booktitle}{\emph{Proceedings of the 55th
  Annual Meeting of the Association for Computational Linguistics (Volume 1:
  Long Papers)}}, Vol.~\bibinfo{volume}{1}. \bibinfo{pages}{873--883}.
\newblock


\bibitem[\protect\citeauthoryear{Ritschel, Aslan, Sedlbauer, and
  Andr{\'e}}{Ritschel et~al\mbox{.}}{2019}]%
        {ritschel2019irony}
\bibfield{author}{\bibinfo{person}{Hannes Ritschel}, \bibinfo{person}{Ilhan
  Aslan}, \bibinfo{person}{David Sedlbauer}, {and} \bibinfo{person}{Elisabeth
  Andr{\'e}}.} \bibinfo{year}{2019}\natexlab{}.
\newblock \showarticletitle{Irony man: augmenting a social robot with the
  ability to use irony in multimodal communication with humans}.
\newblock  (\bibinfo{year}{2019}).
\newblock


\bibitem[\protect\citeauthoryear{Ritschel, Kiderle, Weber, Lingenfelser, Baur,
  and Andr{\'e}}{Ritschel et~al\mbox{.}}{2020}]%
        {ritschel2020multimodal}
\bibfield{author}{\bibinfo{person}{Hannes Ritschel}, \bibinfo{person}{Thomas
  Kiderle}, \bibinfo{person}{Klaus Weber}, \bibinfo{person}{Florian
  Lingenfelser}, \bibinfo{person}{Tobias Baur}, {and}
  \bibinfo{person}{Elisabeth Andr{\'e}}.} \bibinfo{year}{2020}\natexlab{}.
\newblock \showarticletitle{Multimodal joke generation and paralinguistic
  personalization for a socially-aware robot}. In
  \bibinfo{booktitle}{\emph{International Conference on Practical Applications
  of Agents and Multi-Agent Systems}}. Springer, \bibinfo{pages}{278--290}.
\newblock


\bibitem[\protect\citeauthoryear{Sabur, Sari, and Tawami}{Sabur
  et~al\mbox{.}}{2020}]%
        {sabur2020identification}
\bibfield{author}{\bibinfo{person}{Andy~Jefferson Sabur},
  \bibinfo{person}{Retno~Purwani Sari}, {and} \bibinfo{person}{Tatan Tawami}.}
  \bibinfo{year}{2020}\natexlab{}.
\newblock \showarticletitle{Identification Of The Multimodal Structure Of Humor
  In An Animated Superhero Film}. In \bibinfo{booktitle}{\emph{International
  Conference on Language, Linguistics, and Literature (COLALITE) 2020}}.
\newblock


\bibitem[\protect\citeauthoryear{Sane, Tripathi, Sane, and Mamidi}{Sane
  et~al\mbox{.}}{2019}]%
        {sane2019deep}
\bibfield{author}{\bibinfo{person}{Sushmitha~Reddy Sane},
  \bibinfo{person}{Suraj Tripathi}, \bibinfo{person}{Koushik~Reddy Sane}, {and}
  \bibinfo{person}{Radhika Mamidi}.} \bibinfo{year}{2019}\natexlab{}.
\newblock \showarticletitle{Deep learning techniques for humor detection in
  Hindi-English code-mixed tweets}. In \bibinfo{booktitle}{\emph{Proceedings of
  the Tenth Workshop on Computational Approaches to Subjectivity, Sentiment and
  Social Media Analysis}}. \bibinfo{pages}{57--61}.
\newblock


\bibitem[\protect\citeauthoryear{Scheel}{Scheel}{2017}]%
        {scheel2017definitions}
\bibfield{author}{\bibinfo{person}{Tabea Scheel}.}
  \bibinfo{year}{2017}\natexlab{}.
\newblock \showarticletitle{Definitions, theories, and measurement of humor}.
\newblock In \bibinfo{booktitle}{\emph{Humor at work in teams, leadership,
  negotiations, learning and health}}. \bibinfo{publisher}{Springer},
  \bibinfo{pages}{9--29}.
\newblock


\bibitem[\protect\citeauthoryear{Schuller, Valster, Eyben, Cowie, and
  Pantic}{Schuller et~al\mbox{.}}{2012}]%
        {schuller2012avec}
\bibfield{author}{\bibinfo{person}{Bj{\"o}rn Schuller}, \bibinfo{person}{Michel
  Valster}, \bibinfo{person}{Florian Eyben}, \bibinfo{person}{Roddy Cowie},
  {and} \bibinfo{person}{Maja Pantic}.} \bibinfo{year}{2012}\natexlab{}.
\newblock \showarticletitle{Avec 2012: the continuous audio/visual emotion
  challenge}. In \bibinfo{booktitle}{\emph{Proceedings of the 14th ACM
  international conference on Multimodal interaction}}.
  \bibinfo{pages}{449--456}.
\newblock


\bibitem[\protect\citeauthoryear{Song, Williams, Schallert, and Pruitt}{Song
  et~al\mbox{.}}{2021}]%
        {song2021humor}
\bibfield{author}{\bibinfo{person}{Kwangok Song}, \bibinfo{person}{Kyle~M
  Williams}, \bibinfo{person}{Diane~L Schallert}, {and}
  \bibinfo{person}{Alina~Adonyi Pruitt}.} \bibinfo{year}{2021}\natexlab{}.
\newblock \showarticletitle{Humor in multimodal language use: Students’
  Response to a dialogic, social-networking online assignment}.
\newblock \bibinfo{journal}{\emph{Linguistics and Education}}
  \bibinfo{volume}{63} (\bibinfo{year}{2021}), \bibinfo{pages}{100903}.
\newblock


\bibitem[\protect\citeauthoryear{Vasquez and Aslan}{Vasquez and Aslan}{2021}]%
        {vasquez2021cats}
\bibfield{author}{\bibinfo{person}{Camilla Vasquez} {and}
  \bibinfo{person}{Erhan Aslan}.} \bibinfo{year}{2021}\natexlab{}.
\newblock \showarticletitle{“Cats be outside, how about meow”: multimodal
  humor and creativity in an internet meme}.
\newblock \bibinfo{journal}{\emph{Journal of Pragmatics}}
  \bibinfo{volume}{171} (\bibinfo{year}{2021}), \bibinfo{pages}{101--117}.
\newblock


\bibitem[\protect\citeauthoryear{Veronika}{Veronika}{2020}]%
        {veronika2020multimodal}
\bibfield{author}{\bibinfo{person}{Zhigailova Veronika}.}
  \bibinfo{year}{2020}\natexlab{}.
\newblock \showarticletitle{Multimodal Discourse Analysis of Humor in Picture
  Books for Children}.
\newblock  (\bibinfo{year}{2020}).
\newblock


\bibitem[\protect\citeauthoryear{Warren, Barsky, and McGraw}{Warren
  et~al\mbox{.}}{2018}]%
        {warren2018humor}
\bibfield{author}{\bibinfo{person}{Caleb Warren}, \bibinfo{person}{Adam
  Barsky}, {and} \bibinfo{person}{A~Peter McGraw}.}
  \bibinfo{year}{2018}\natexlab{}.
\newblock \showarticletitle{Humor, comedy, and consumer behavior}.
\newblock \bibinfo{journal}{\emph{Journal of Consumer Research}}
  \bibinfo{volume}{45}, \bibinfo{number}{3} (\bibinfo{year}{2018}),
  \bibinfo{pages}{529--552}.
\newblock


\bibitem[\protect\citeauthoryear{Weller and Seppi}{Weller and Seppi}{2019}]%
        {weller2019humor}
\bibfield{author}{\bibinfo{person}{Orion Weller} {and} \bibinfo{person}{Kevin
  Seppi}.} \bibinfo{year}{2019}\natexlab{}.
\newblock \showarticletitle{Humor detection: A transformer gets the last
  laugh}.
\newblock \bibinfo{journal}{\emph{arXiv preprint arXiv:1909.00252}}
  (\bibinfo{year}{2019}).
\newblock


\bibitem[\protect\citeauthoryear{Xie, Girshick, Doll{\'a}r, Tu, and He}{Xie
  et~al\mbox{.}}{2017}]%
        {xie2017aggregated}
\bibfield{author}{\bibinfo{person}{Saining Xie}, \bibinfo{person}{Ross
  Girshick}, \bibinfo{person}{Piotr Doll{\'a}r}, \bibinfo{person}{Zhuowen Tu},
  {and} \bibinfo{person}{Kaiming He}.} \bibinfo{year}{2017}\natexlab{}.
\newblock \showarticletitle{Aggregated residual transformations for deep neural
  networks}. In \bibinfo{booktitle}{\emph{Proceedings of the IEEE conference on
  computer vision and pattern recognition}}. \bibinfo{pages}{1492--1500}.
\newblock


\bibitem[\protect\citeauthoryear{Yang, Ai, and Hirschberg}{Yang
  et~al\mbox{.}}{2019}]%
        {yang2019multimodal}
\bibfield{author}{\bibinfo{person}{Zixiaofan Yang}, \bibinfo{person}{Lin Ai},
  {and} \bibinfo{person}{Julia Hirschberg}.} \bibinfo{year}{2019}\natexlab{}.
\newblock \showarticletitle{Multimodal Indicators of Humor in Videos}. In
  \bibinfo{booktitle}{\emph{2019 IEEE Conference on Multimedia Information
  Processing and Retrieval (MIPR)}}. IEEE, \bibinfo{pages}{538--543}.
\newblock


\bibitem[\protect\citeauthoryear{Zadeh, Liang, Poria, Cambria, and
  Morency}{Zadeh et~al\mbox{.}}{2018}]%
        {zadeh2018acl}
\bibfield{author}{\bibinfo{person}{Amir Zadeh}, \bibinfo{person}{Paul~Pu
  Liang}, \bibinfo{person}{Soujanya Poria}, \bibinfo{person}{Erik Cambria},
  {and} \bibinfo{person}{Louis-Philippe Morency}.}
  \bibinfo{year}{2018}\natexlab{}.
\newblock \showarticletitle{Multimodal Language Analysis in the Wild: CMU-MOSEI
  Dataset and Interpretable Dynamic Fusion Graph}. In
  \bibinfo{booktitle}{\emph{Proceedings of the 56th Annual Meeting of the
  Association for Computational Linguistics (Volume 1: Long Papers)}}
  (Melbourne, Australia). \bibinfo{publisher}{Association for Computational
  Linguistics}, \bibinfo{pages}{2236--2246}.
\newblock
\urldef\tempurl%
\url{http://aclweb.org/anthology/P18-1208}
\showURL{%
\tempurl}


\bibitem[\protect\citeauthoryear{Zadeh, Zellers, Pincus, and Morency}{Zadeh
  et~al\mbox{.}}{2016}]%
        {zadeh:mosi}
\bibfield{author}{\bibinfo{person}{A. Zadeh}, \bibinfo{person}{R. Zellers},
  \bibinfo{person}{E. Pincus}, {and} \bibinfo{person}{L.~P. Morency}.}
  \bibinfo{year}{2016}\natexlab{}.
\newblock \showarticletitle{{Multimodal Sentiment Intensity Analysis in Videos:
  Facial Gestures and Verbal Messages}}.
\newblock \bibinfo{journal}{\emph{IEEE Intelligent Systems}}
  \bibinfo{volume}{31}, \bibinfo{number}{6} (\bibinfo{date}{Nov}
  \bibinfo{year}{2016}), \bibinfo{pages}{82--88}.
\newblock
\showISSN{1541-1672}
\urldef\tempurl%
\url{https://doi.org/10.1109/MIS.2016.94}
\showDOI{\tempurl}


\bibitem[\protect\citeauthoryear{Zhao, Cattle, Papalexakis, and Ma}{Zhao
  et~al\mbox{.}}{2019}]%
        {zhao2019embedding}
\bibfield{author}{\bibinfo{person}{Zhenjie Zhao}, \bibinfo{person}{Andrew
  Cattle}, \bibinfo{person}{Evangelos Papalexakis}, {and}
  \bibinfo{person}{Xiaojuan Ma}.} \bibinfo{year}{2019}\natexlab{}.
\newblock \showarticletitle{Embedding lexical features via tensor decomposition
  for small sample humor recognition}. In \bibinfo{booktitle}{\emph{Proceedings
  of the 2019 Conference on Empirical Methods in Natural Language Processing
  and the 9th International Joint Conference on Natural Language Processing
  (EMNLP-IJCNLP)}}.
\newblock


\end{thebibliography}

\end{document}


\title{Toward Developing a Multimodal Multi-party Hindi Humorous Dataset for Humor Recognition in Conversations}

\renewcommand{\shortauthors}{Trovato and Tobin, et al.}


\maketitle

\appendix

\section{Error Analysis}
For the global audience, we also translate these Hindi utterances into English (c.f. Table \ref{tab-error_analysis_appendix}).

\begin{table}[!h]
\small
\centering
\resizebox{0.49\textwidth}{!}
{
\begin{tabular}{|l|p{18em}||c||c|c| }
\hline

\multicolumn{5}{|c|}{\bf \em Correct Prediction} \\ \hline
& & & \multicolumn{2}{c|}{\bf \em Predicted} \\ \cline{4-5}

& \bf  & \bf  & \bf Unimodal & \bf Multimodal \\ 
& \bf Utterances & \bf Actual & \bf (T) & \bf (T+V+A) \\ \hline \hline

1 & Oh in old age I would have taken you myself. & humor & \textcolor{red}{non-humor} & humor \\ \hline

2 & This Dilruba says that he was the only son of Chhotumal Motumal Crorepati in his previous life, right? & non-humor & \textcolor{red}{humor} & non-humor \\ \hline

3 & Staff guys! I mean & humor & \textcolor{red}{non-humor} & humor \\ \hline

4 & remember something? remember something? & non-humor & \textcolor{red}{humor} & non-humor \\ \hline

5 & Do not do such inauspicious things while leaving. & humor & \textcolor{red}{non-humor} & humor  \\ \hline \hline

\multicolumn{5}{|c|}{\bf \em Incorrect Prediction} \\ \hline

1 & these guys are killing me a lot & humor & \textcolor{red}{non-humor} & \textcolor{red}{non-humor} \\ \hline

2 & hey, Toto! & non-humor & non-humor & \textcolor{red}{humor} \\ \hline

3 & Pinky bought this & humor & humor & \textcolor{red}{non-humor} \\ \hline

4 & Hey brother, my original gun is left here & non-humor & \textcolor{red}{humor} & \textcolor{red}{humor} \\ \hline

5 & which is real which is fake & non-humor & non-humor & \textcolor{red}{humor}  \\ \hline

\end{tabular}
}
\caption{Some correctly and incorrectly predicted samples}
\label{tab-error_analysis_appendix}
\end{table}